\documentclass{book}

\usepackage[utf8]{inputenc}
\usepackage{pdfpages}
\usepackage{fancyhdr}
\usepackage{hyperref}

\usepackage{combelow}
\usepackage{newunicodechar}

\usepackage{ amstext, epstopdf, booktabs, verbatim, gensymb, geometry, lmodern}
\usepackage{tocloft}
\newlength{\standardchapnumwidth}
\AtBeginDocument{\setlength{\standardchapnumwidth}{\cftchapnumwidth}}

\newunicodechar{Ș}{\cb{S}}

\newcommand*\cpiType{Volume 2}
\newcommand*\Date{June 2021}
\newcommand*\Author{Federico Castagna\\ Francesca Mosca\\ Jack Mumford\\ \cb{S}tefan Sarkadi \\Andreas Xydis}

\definecolor{myblue}{HTML}{154360}
\definecolor{emerald}{HTML}{3cb371}

\begin{document}

\newgeometry{margin = 0in}

\pagecolor{emerald}
\setlength{\fboxsep}{0pt}
\hfill \colorbox{myblue}{\makebox[3.22in][r]{\shortstack[r]{\vspace{3.3in}}}}%
\setlength{\fboxsep}{15pt}
\setlength{\fboxrule}{5pt}
\colorbox{white}{\makebox[\linewidth][c]{\includegraphics[width=1.3in]{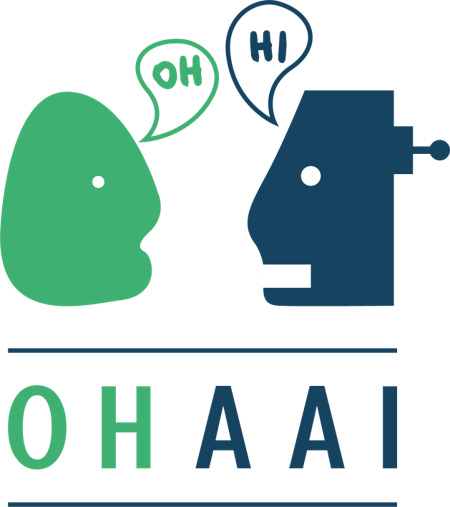}\hspace{0.35in} \shortstack[l]{\vspace{10pt}\fontsize{40}{40}\rmfamily\color{myblue} Online Handbook of \\
\vspace{10pt}
\fontsize{40}{40}\rmfamily\color{myblue} Argumentation for AI\\%
\fontsize{20}{20}\rmfamily\color{myblue} \cpiType}}}%
\setlength{\fboxsep}{0pt}
\vspace{-0.25pt}
\hfill \colorbox{myblue}{\hspace{.25in} \parbox{2.97in}{\vspace{2in} \color{white} \large{Edited by \\ \\ \Author  \\ \\  \Date \vspace{2.15in} \vfill}}}%
\restoregeometry

\nopagecolor

\thispagestyle{empty}
\pagenumbering{gobble}

\begin{center}
    \textbf{{\huge Preface}}
\end{center}

\hfill

This volume contains revised versions of the papers selected for the second volume of the Online Handbook of Argumentation for AI (OHAAI). Previously, formal theories of argument and argument interaction have been proposed and studied, and this has led to the more recent study of computational models of argument. Argumentation, as a field within artificial intelligence (AI), is highly relevant for researchers interested in symbolic representations of knowledge and defeasible reasoning. 
The purpose of this handbook is to provide an open access and curated anthology for the argumentation research community. OHAAI is designed to serve as a research hub to keep track of the latest and upcoming PhD-driven research on the theory and application of argumentation in all areas related to AI. The handbook’s goals are to:

\begin{enumerate}
    \item Encourage collaboration and knowledge discovery between members of the argumentation community.
    \item Provide a platform for PhD students to have their work published in a citable peer-reviewed venue.
    \item Present an indication of the scope and quality of PhD research involving argumentation for AI.
\end{enumerate}

The papers in this volume are those selected for inclusion in OHAAI Vol.2 following a back-and-forth peer-review process. The volume thus presents a strong representation of the contemporary state of the art research of argumentation in AI that has been strictly undertaken during PhD studies. Papers in this volume are listed alphabetically by author. We hope that you will enjoy reading this handbook.
\begin{flushright}
\noindent\begin{tabular}{r}
\makebox[1.3in]{}\\
\textit{Editors}\\
Federico Castagna\\
Francesca Mosca\\
Jack Mumford\\
\cb{S}tefan Sarkadi\\
Andreas Xydis\\\\
\textbf{June 2021}
\end{tabular}
\end{flushright}


\pagenumbering{gobble}

\begin{center}
    \textbf{{\huge Acknowledgements}}
\end{center}

\hfill

\noindent
We thank the senior researchers in the area of Argumentation and Artificial Intelligence for their efforts in spreading the word about the OHAAI project with early-career researchers.

\hfill

\noindent
We are especially thankful to Henry Prakken and the COMMA 2020 organisers for providing us the opportunity to promote the OHAAI project at their conference.

\hfill

\noindent
We are also grateful to ArXiv for their willingness to publish this handbook.

\hfill

\noindent
Our sincere gratitude to Costanza Hardouin for her fantastic work in designing the OHAAI logo.

\hfill

\noindent
We owe many thanks to Sanjay Modgil for helping to form the motivation for the handbook, and to Elizabeth Black and Simon Parsons for their advice and guidance that enabled the OHAAI project to come to fruition.

\hfill

\noindent
Special thanks must go to the contributing authors: Andreas Br{\"a}nnstr{\"o}m, Federico Castagna, Théo Duchatelle, Matt Foulis, Timotheus Kampik, Isabelle Kuhlmann, Lars Malmqvist, Mariela Morveli-Espinoza, Jack Mumford, Stipe Pand\v zi\' c, Robin Schaefer, Luke Thorburn, Andreas Xydis, Antonio Yuste-Ginel, and Heng Zheng. Thank you for making the world of argumentation greater! 

\newgeometry{margin = 0.9in}

\pagenumbering{arabic}

\tableofcontents 
\thispagestyle{empty}
\clearpage




\pagestyle{fancy}
\addtocontents{toc}{\protect\renewcommand{\protect\cftchapleader}
     {\protect\cftdotfill{\cftsecdotsep}}}
\addtocontents{toc}{\setlength{\protect\cftchapnumwidth}{0pt}}

\refstepcounter{chapter}\label{1}
\addcontentsline{toc}{chapter}{Strategic Argumentation to deal with Interactions between Intelligent Systems and Humans \\ \textnormal{\textit{Andreas Brännström}}}
\includepdf[pages=-,pagecommand={\thispagestyle{plain}}]{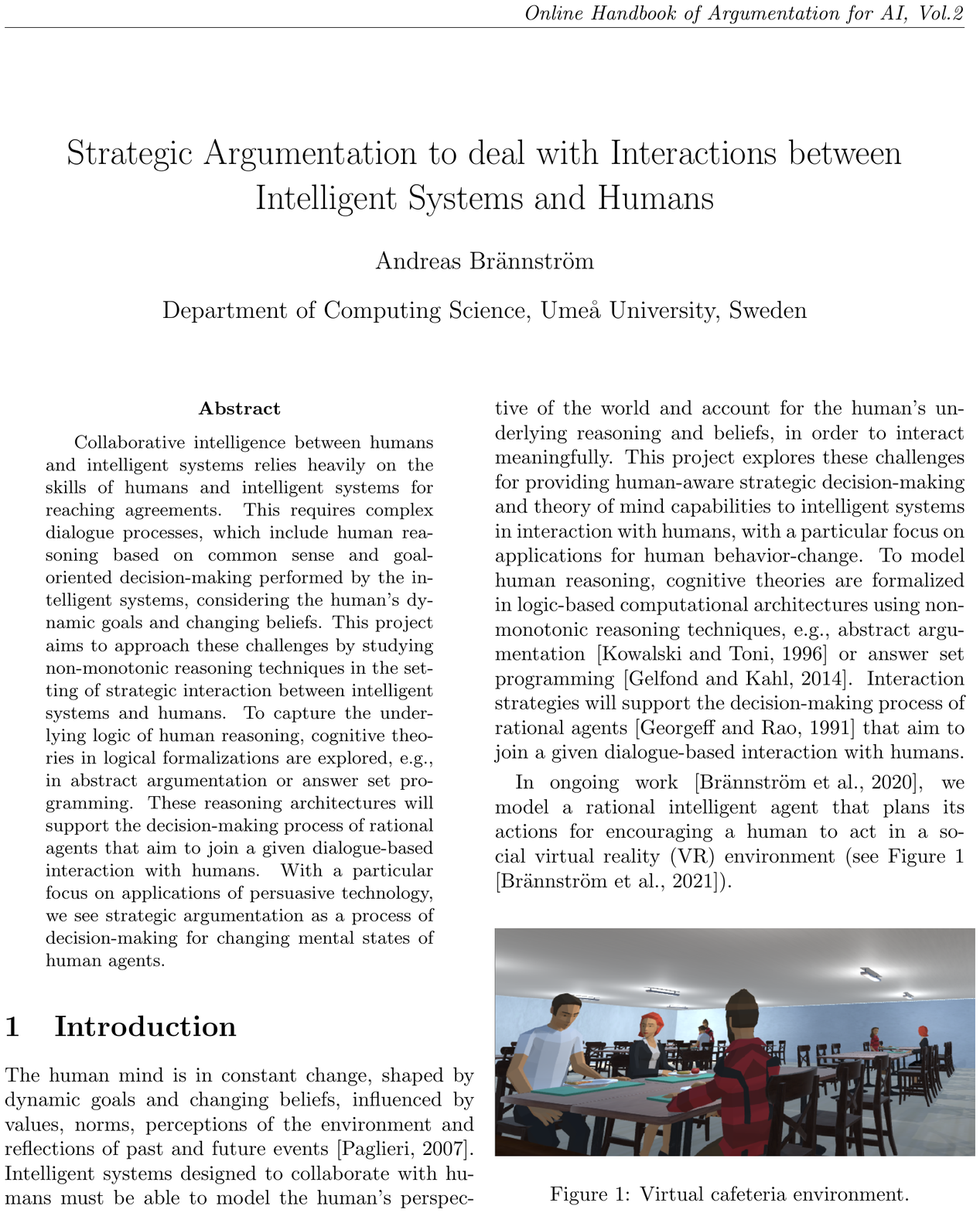}

\refstepcounter{chapter}\label{2}
\addcontentsline{toc}{chapter}{Labelling procedures for Dialectical Classical Logic Argumentation\\
\textnormal{\textit{Federico Castagna}}}
\includepdf[pages=-,pagecommand={\thispagestyle{plain}}]{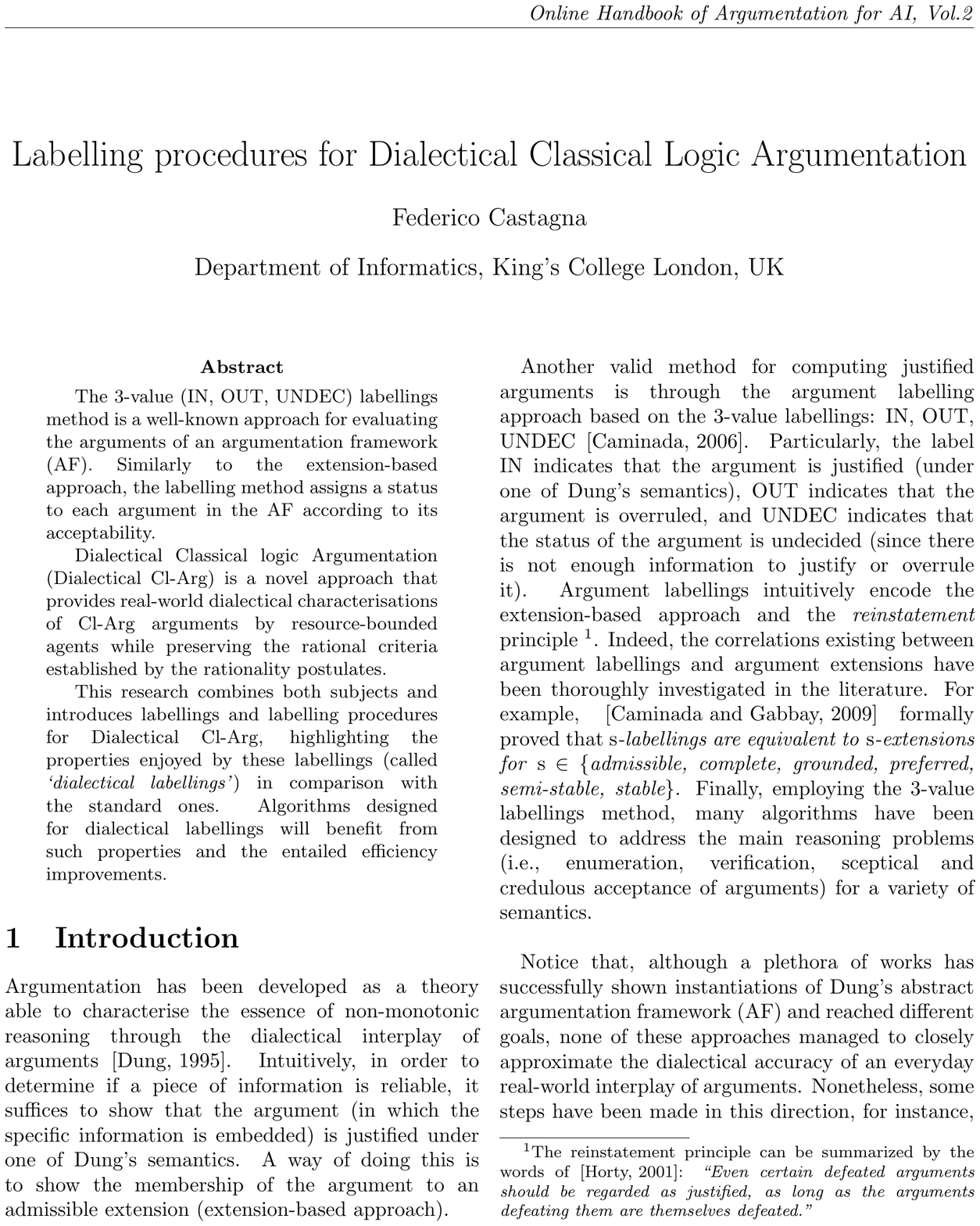}

\refstepcounter{chapter}\label{3}
\addcontentsline{toc}{chapter}{Towards a Generic Logical Encoding for Argumentation \\
\textnormal{\textit{Théo Duchatelle}}}
\includepdf[pages=-,pagecommand={\thispagestyle{plain}}]{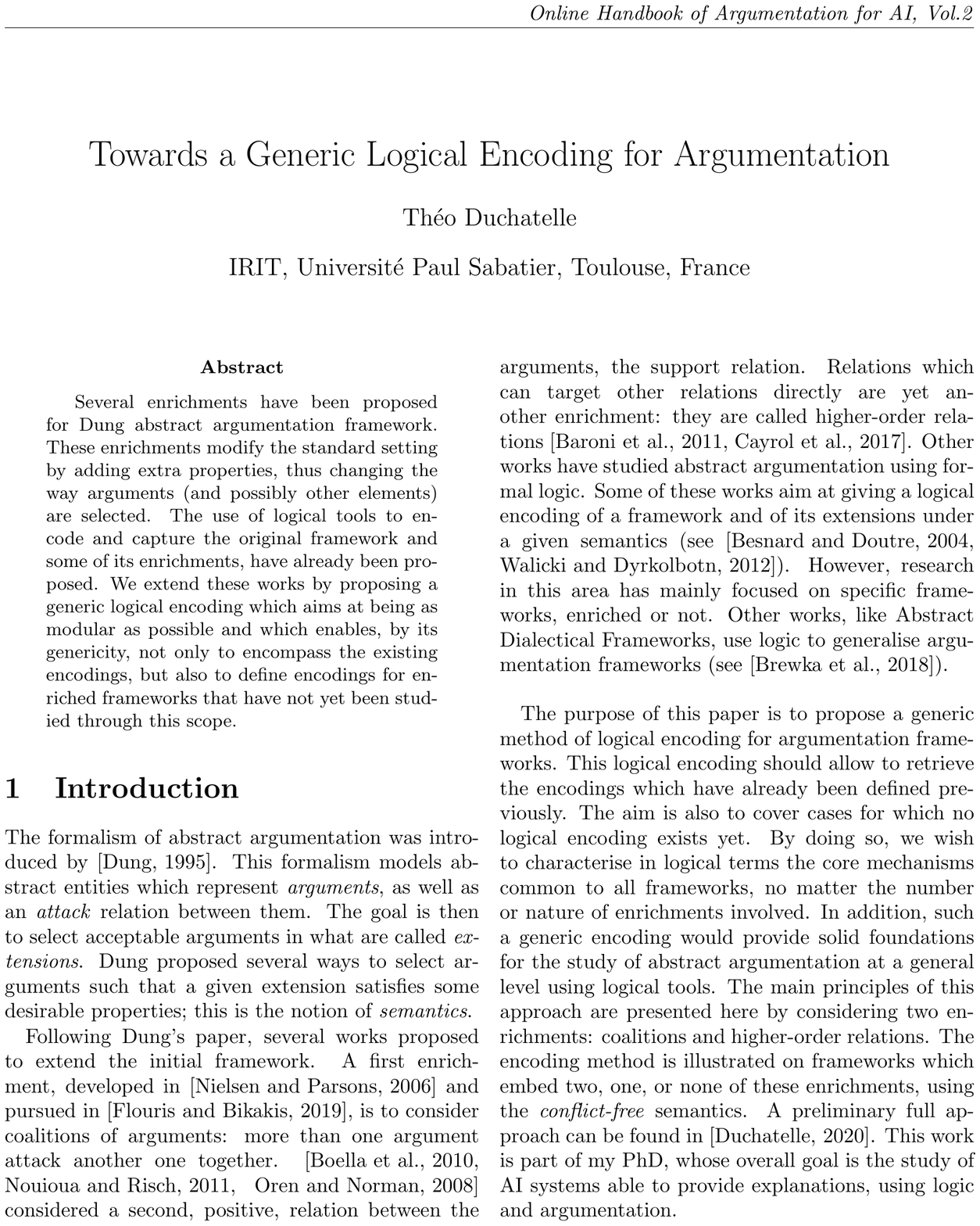}

\refstepcounter{chapter}\label{4}
\addcontentsline{toc}{chapter}{Computational Approaches to Fallacy Detection in Natural Language Arguments \\
\textnormal{\textit{Matt Foulis}}}
\includepdf[pages=-,pagecommand={\thispagestyle{plain}}]{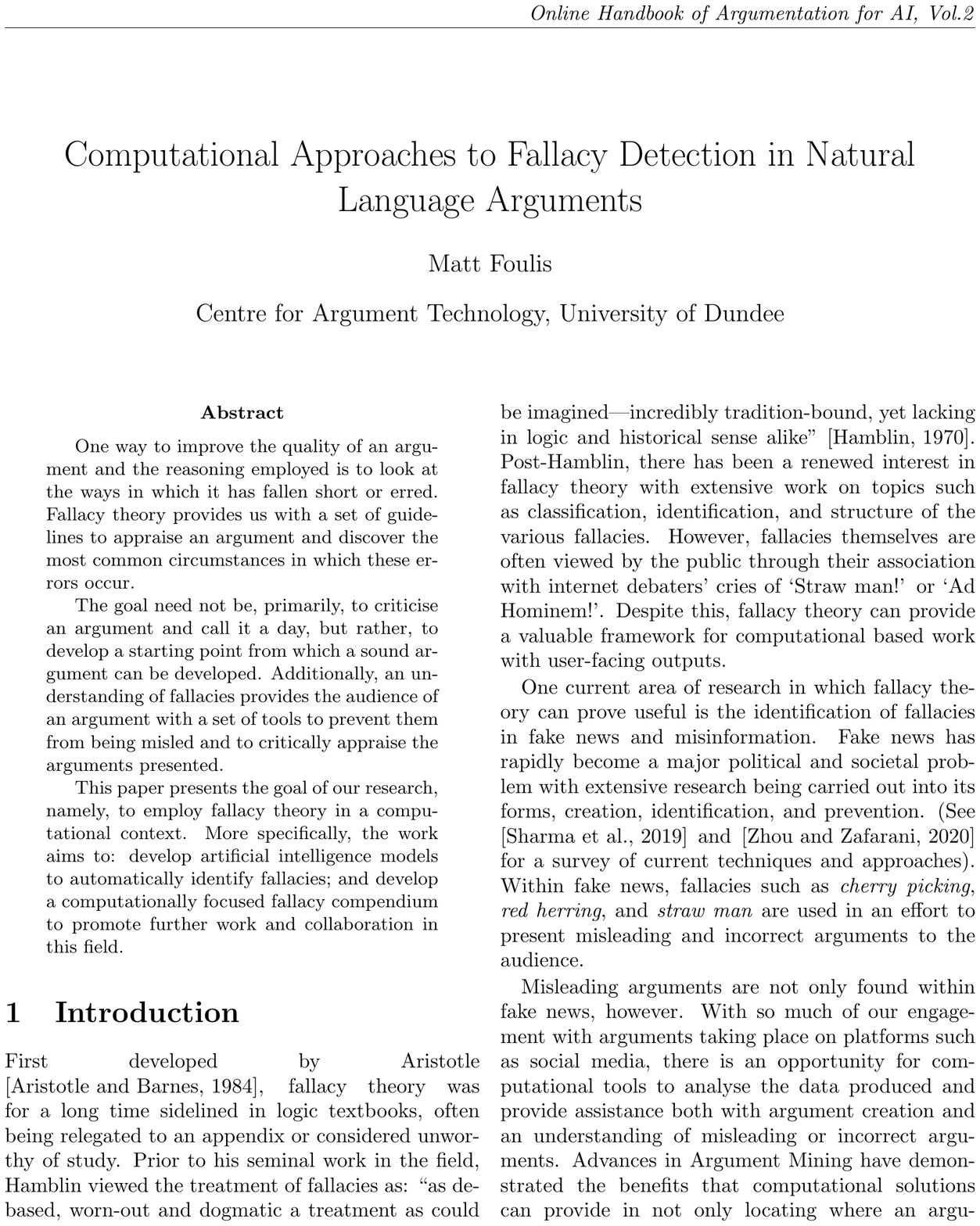}

\refstepcounter{chapter}\label{5}
\addcontentsline{toc}{chapter}{Consistency Principles for Sequential Abstract Argumentation \\
\textnormal{\textit{Timotheus Kampik}}}
\includepdf[pages=-,pagecommand={\thispagestyle{plain}}]{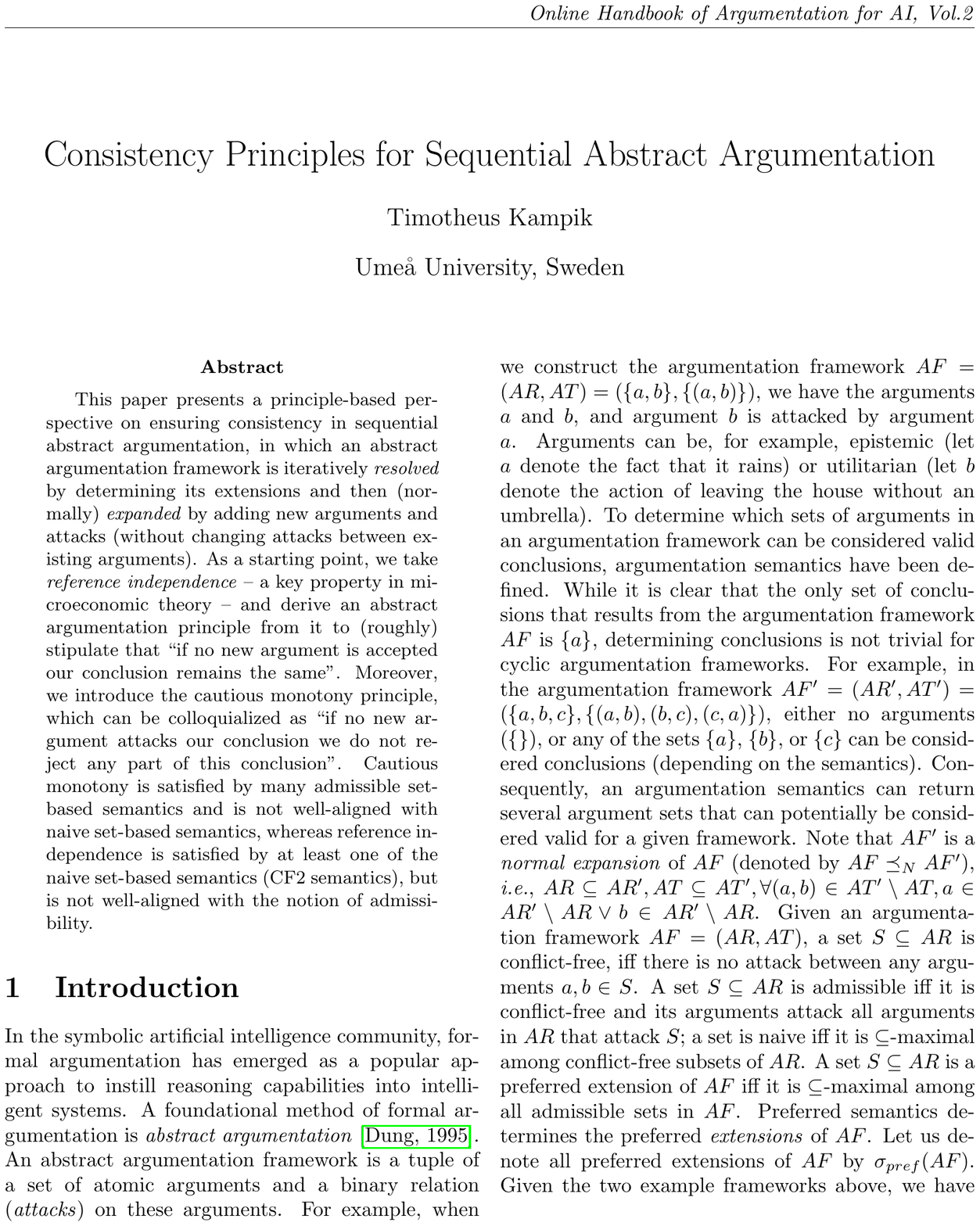}

\refstepcounter{chapter}\label{6}
\addcontentsline{toc}{chapter}{Towards Eliciting Attacks in Abstract Argumentation Frameworks \\
\textnormal{\textit{Isabelle Kuhlmann}}}
\includepdf[pages=-,pagecommand={\thispagestyle{plain}}]{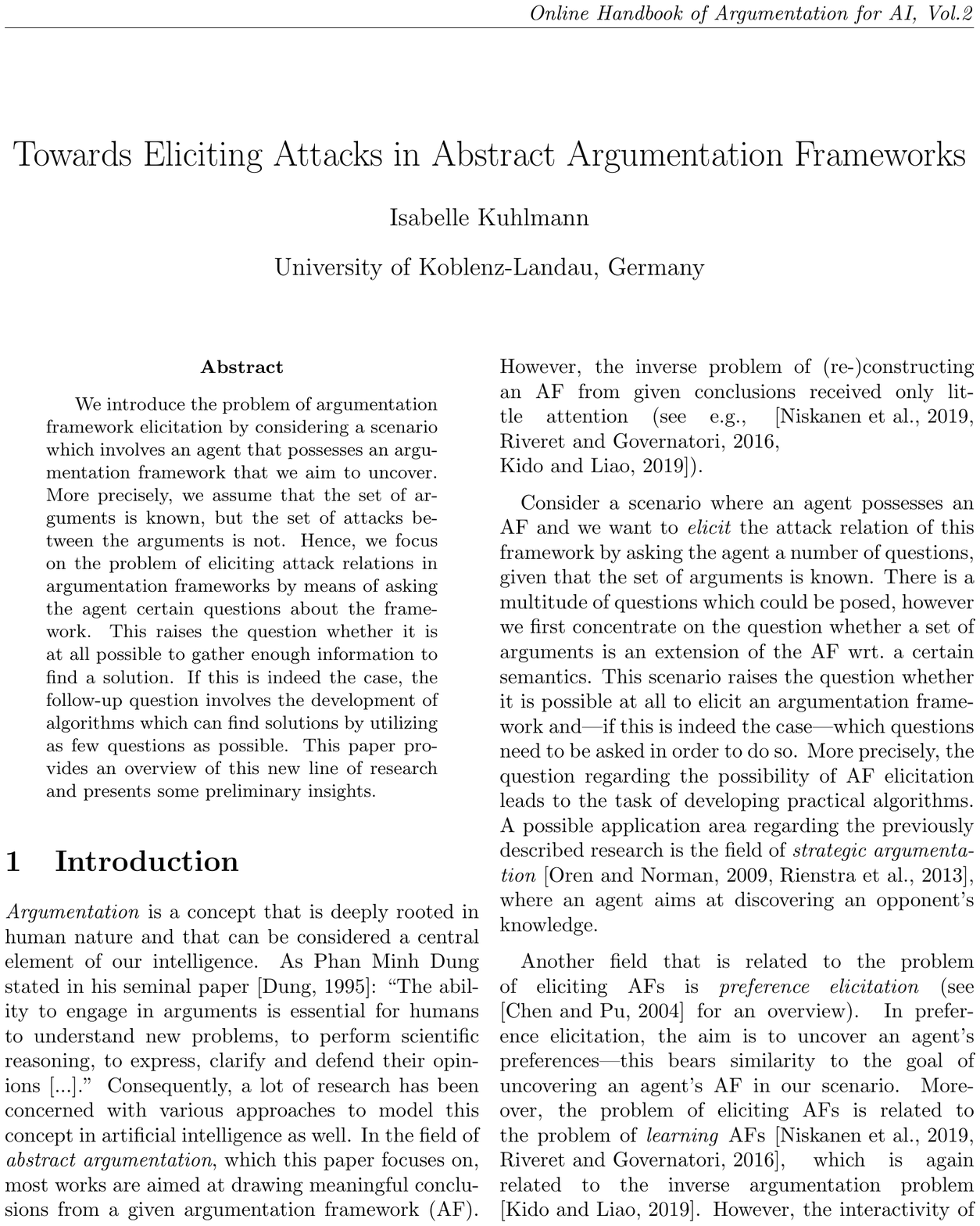}

\refstepcounter{chapter}\label{7}
\addcontentsline{toc}{chapter}{Approximate Solutions to Argumentation Frameworks with Graph Neural Networks  \\
\textnormal{\textit{Lars Malmqvist}}}
\includepdf[pages=-,pagecommand={\thispagestyle{plain}}]{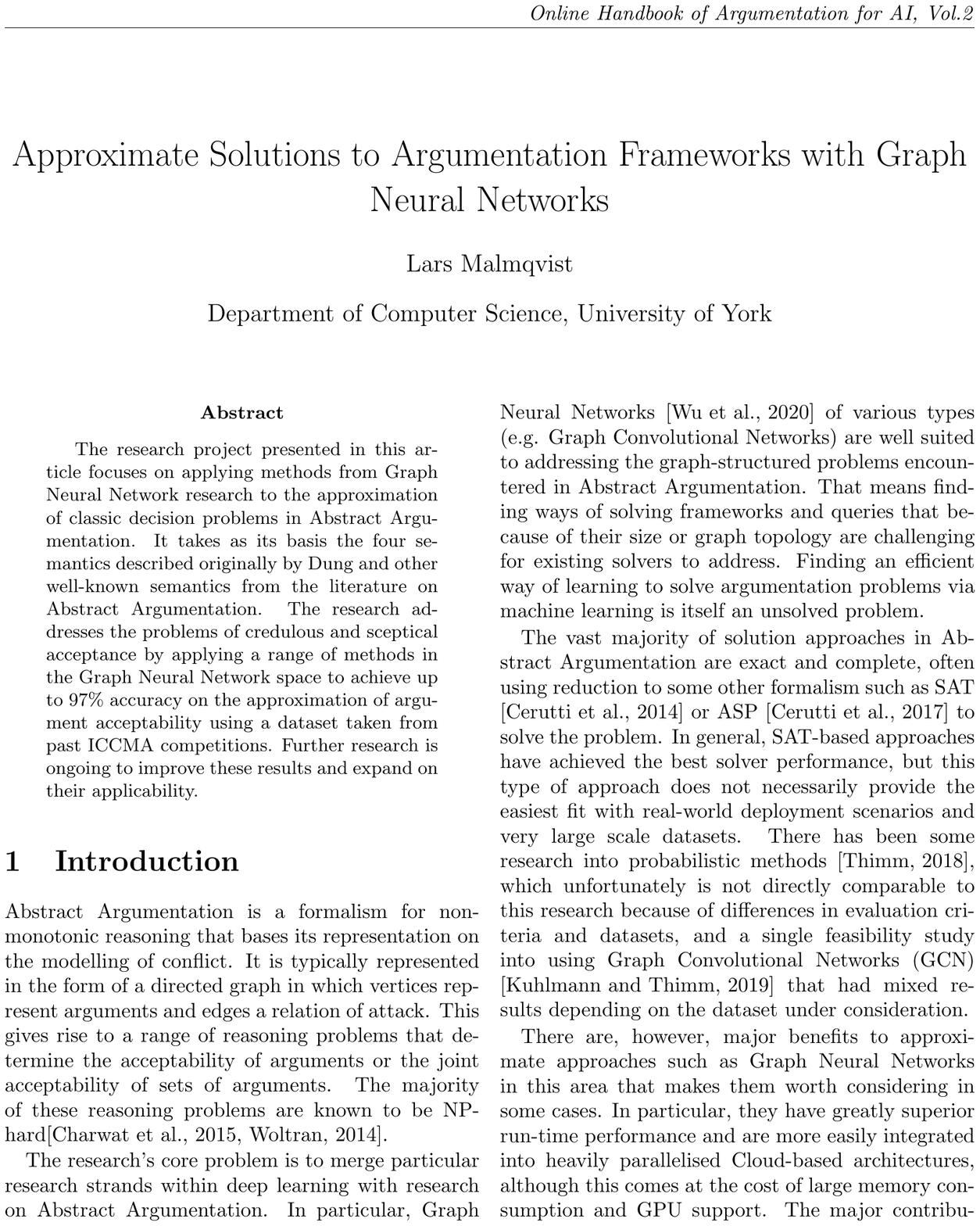}

\refstepcounter{chapter}\label{8}
\addcontentsline{toc}{chapter}{On the Calculation of Rhetorical Arguments Strength for
Persuasive Negotiation in Multi-agent Systems \\
\textnormal{\textit{Mariela Morveli-Espinoza}}}
\includepdf[pages=-,pagecommand={\thispagestyle{plain}}]{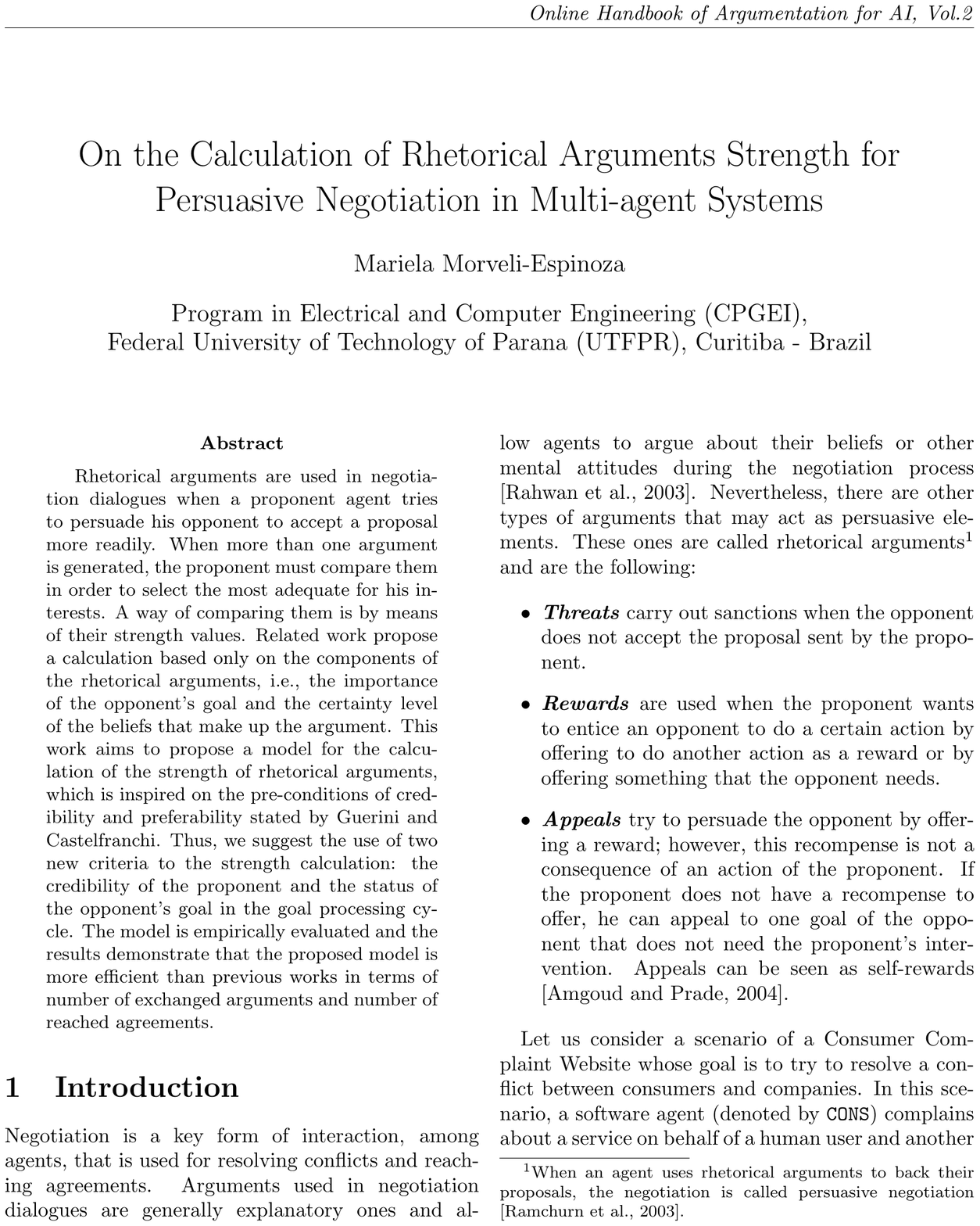}

\refstepcounter{chapter}\label{9}
\addcontentsline{toc}{chapter}{Learning Defeat Relations from Data Using Neural Argumentation Networks (NANs) \\
\textnormal{\textit{Jack Mumford}}}
\includepdf[pages=-,pagecommand={\thispagestyle{plain}}]{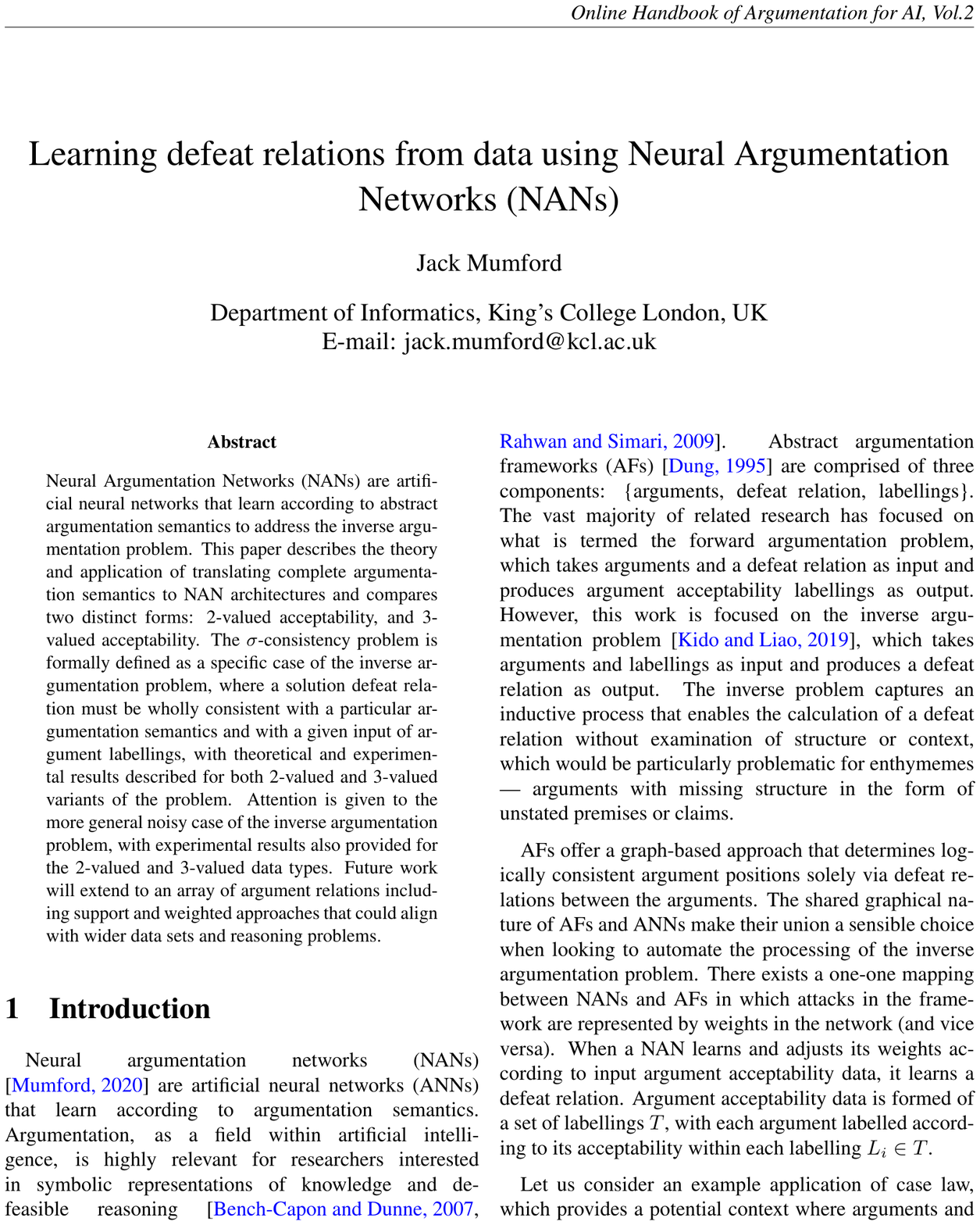}

\refstepcounter{chapter}\label{10}
\addcontentsline{toc}{chapter}{From the Logic of Proofs to the Logic of Arguments \\
\textnormal{\textit{Stipe Pand\v zi\'c }}}
\includepdf[pages=-,pagecommand={\thispagestyle{plain}}]{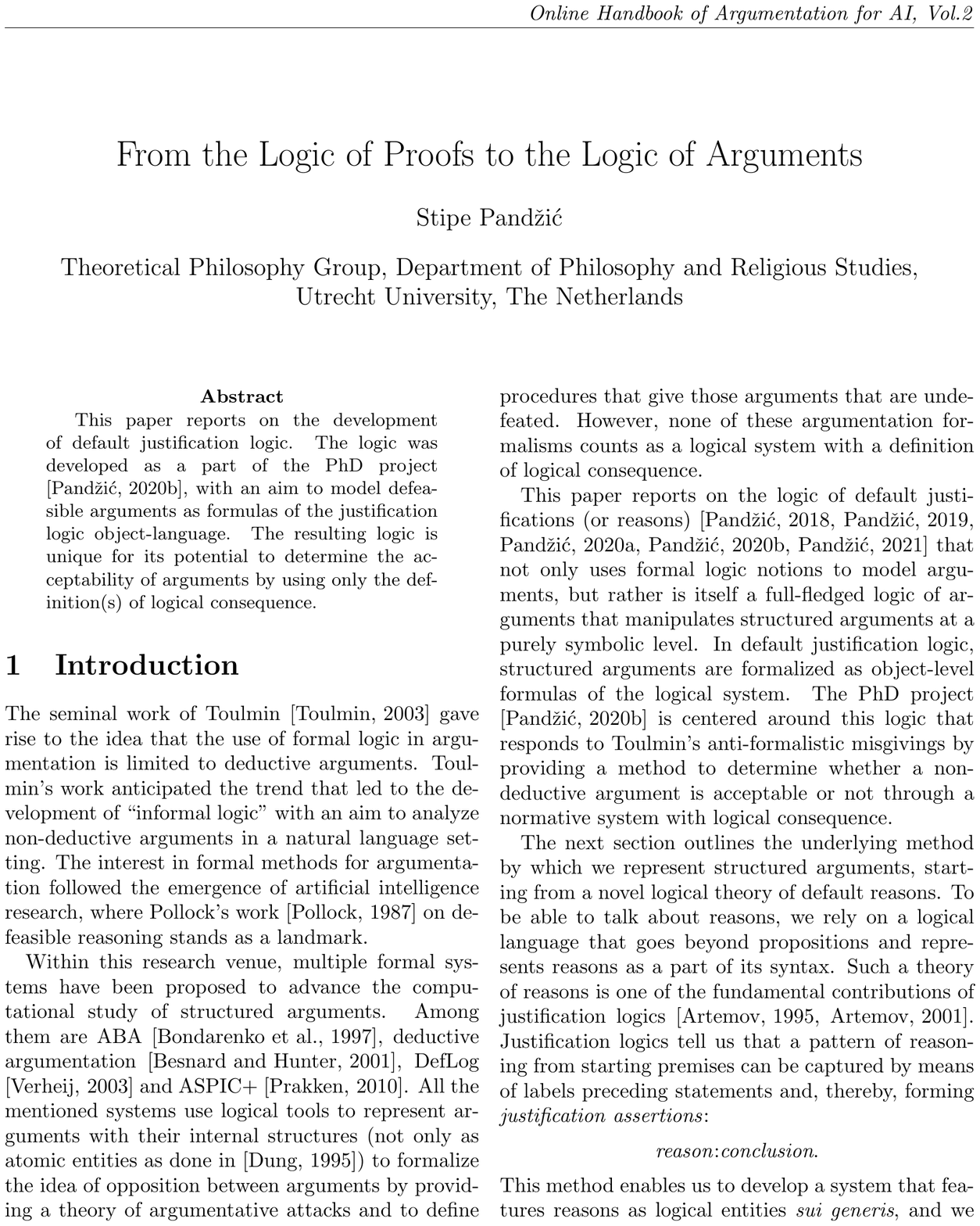}

\refstepcounter{chapter}\label{11}
\addcontentsline{toc}{chapter}{Building an Argument Mining Pipeline for Tweets \\
\textnormal{\textit{Robin Schaefer}}}
\includepdf[pages=-,pagecommand={\thispagestyle{plain}}]{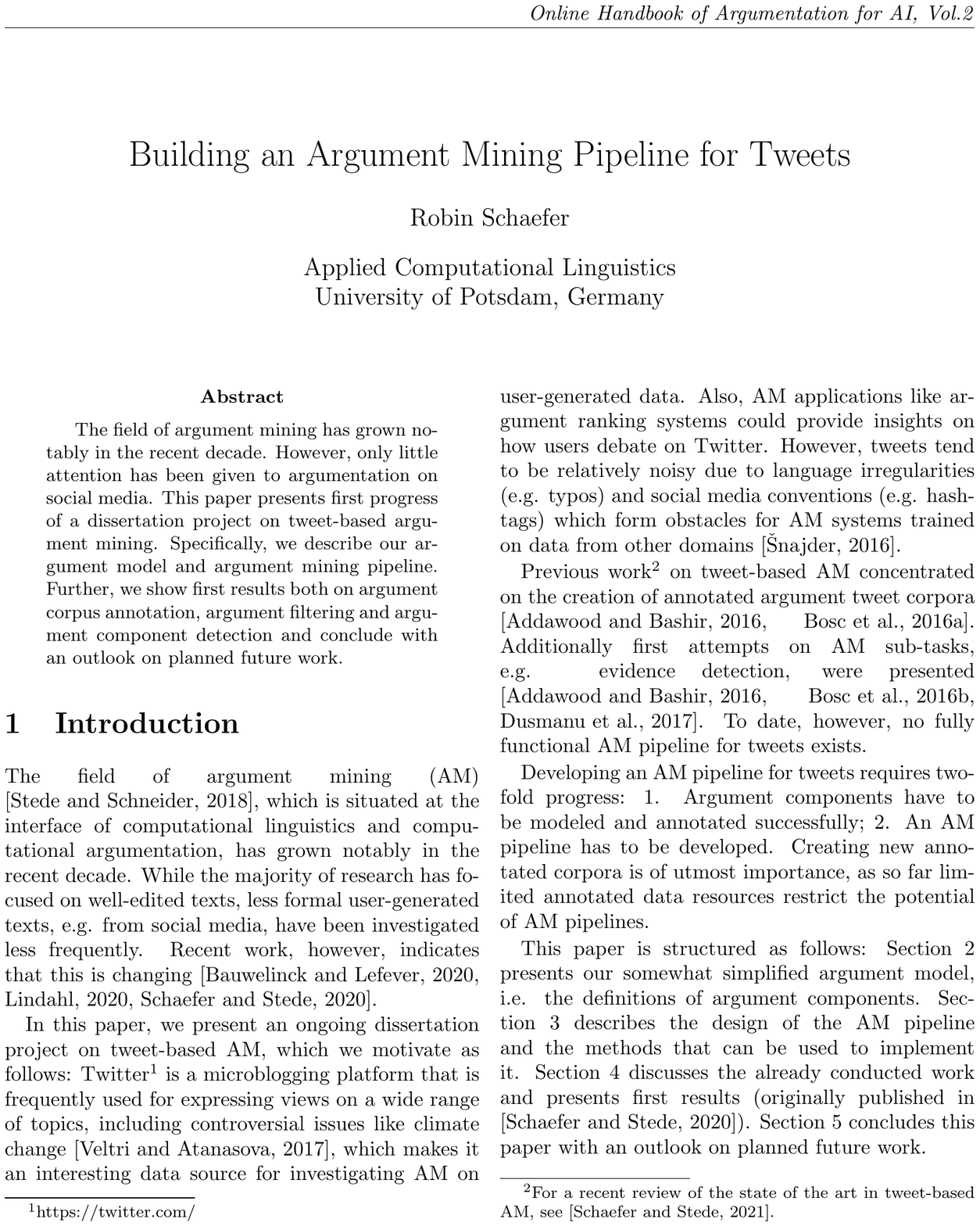}

\refstepcounter{chapter}\label{12}
\addcontentsline{toc}{chapter}{Towards Argument Embeddings \\
\textnormal{\textit{Luke Thorburn}}}
\includepdf[pages=-,pagecommand={\thispagestyle{plain}}]{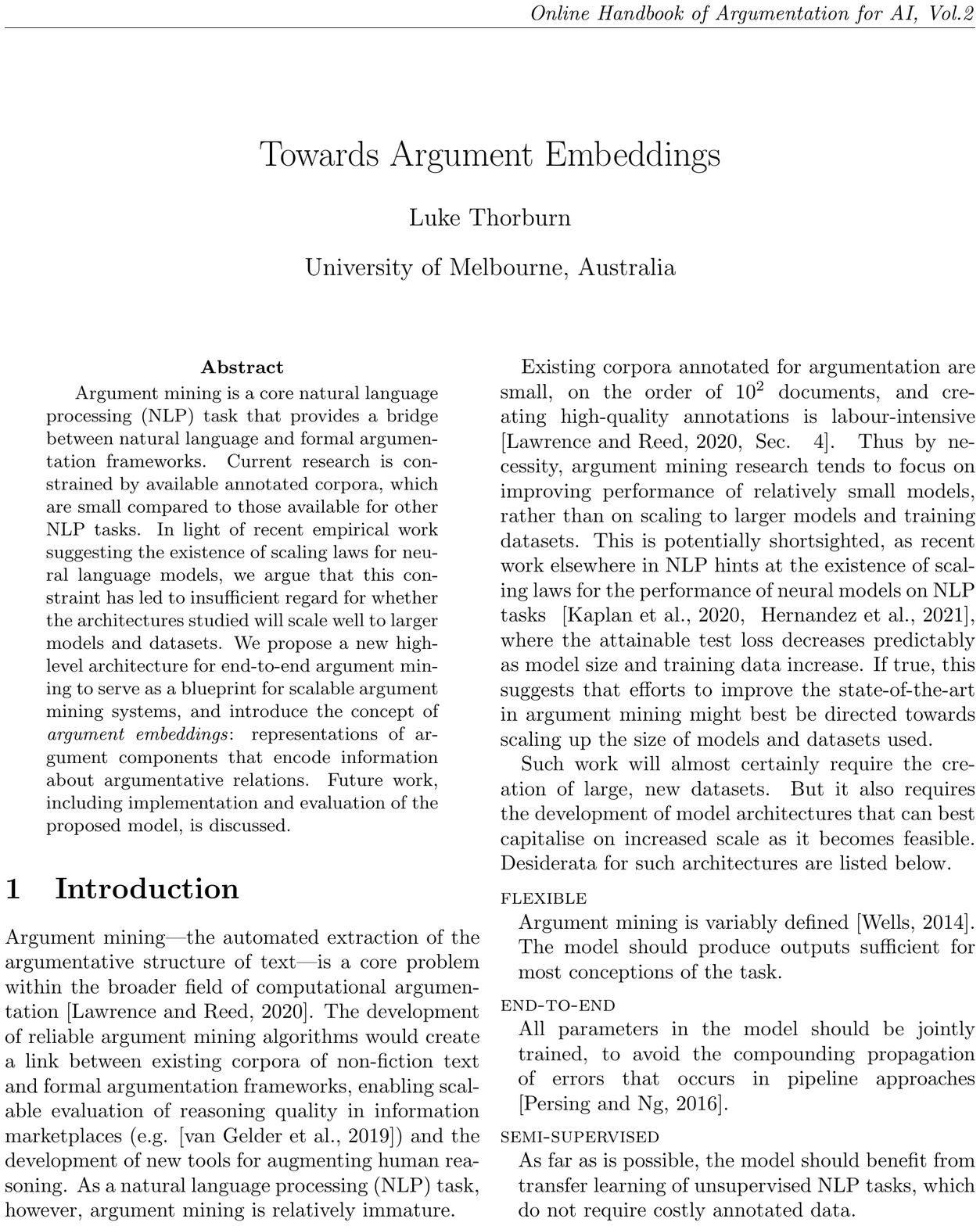}

\refstepcounter{chapter}\label{13}
\addcontentsline{toc}{chapter}{A Dialogue Framework for Enthymemes \\
\textnormal{\textit{Andreas Xydis}}}
\includepdf[pages=-,pagecommand={\thispagestyle{plain}}]{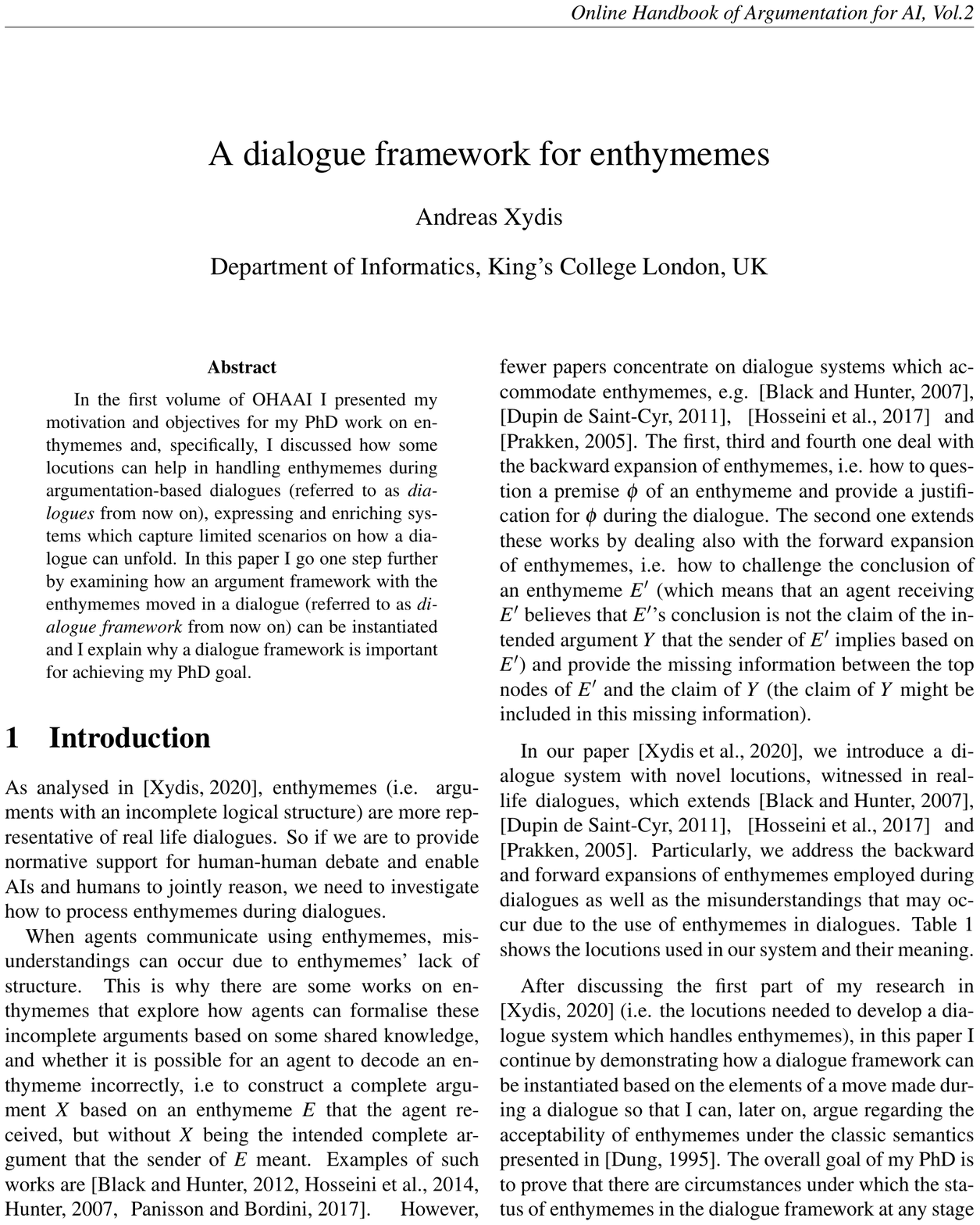}

\refstepcounter{chapter}\label{14}
\addcontentsline{toc}{chapter}{On Some Formal Relations between Arguing and Believing\\
\textnormal{\textit{Antonio Yuste-Ginel}}}
\includepdf[pages=-,pagecommand={\thispagestyle{plain}}]{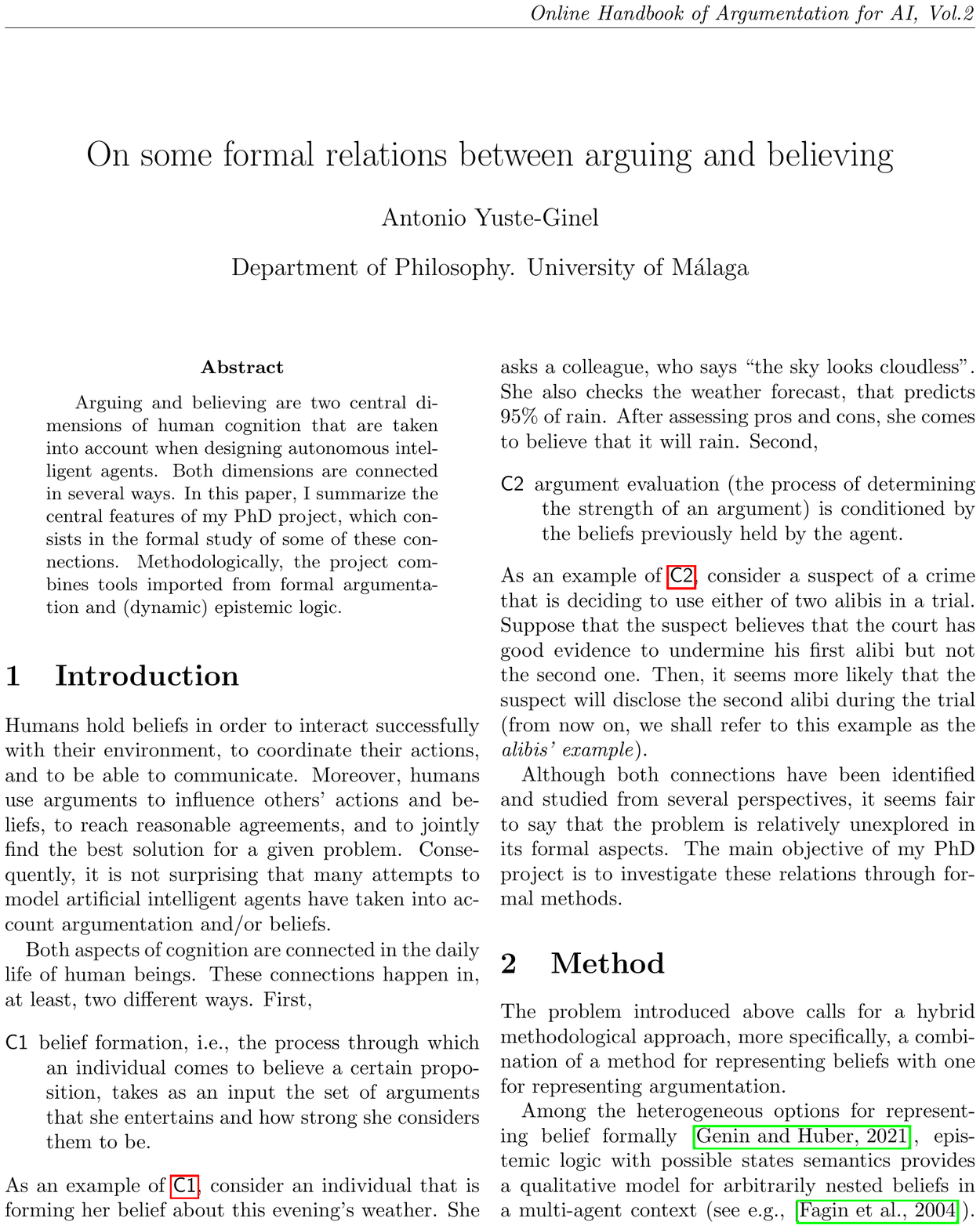}

\refstepcounter{chapter}\label{15}
\addcontentsline{toc}{chapter}{Modeling Case-Based Reasoning with the Precedent Model Formalism: An Overview \\
\textnormal{\textit{Heng Zheng}}}
\includepdf[pages=-,pagecommand={\thispagestyle{plain}}]{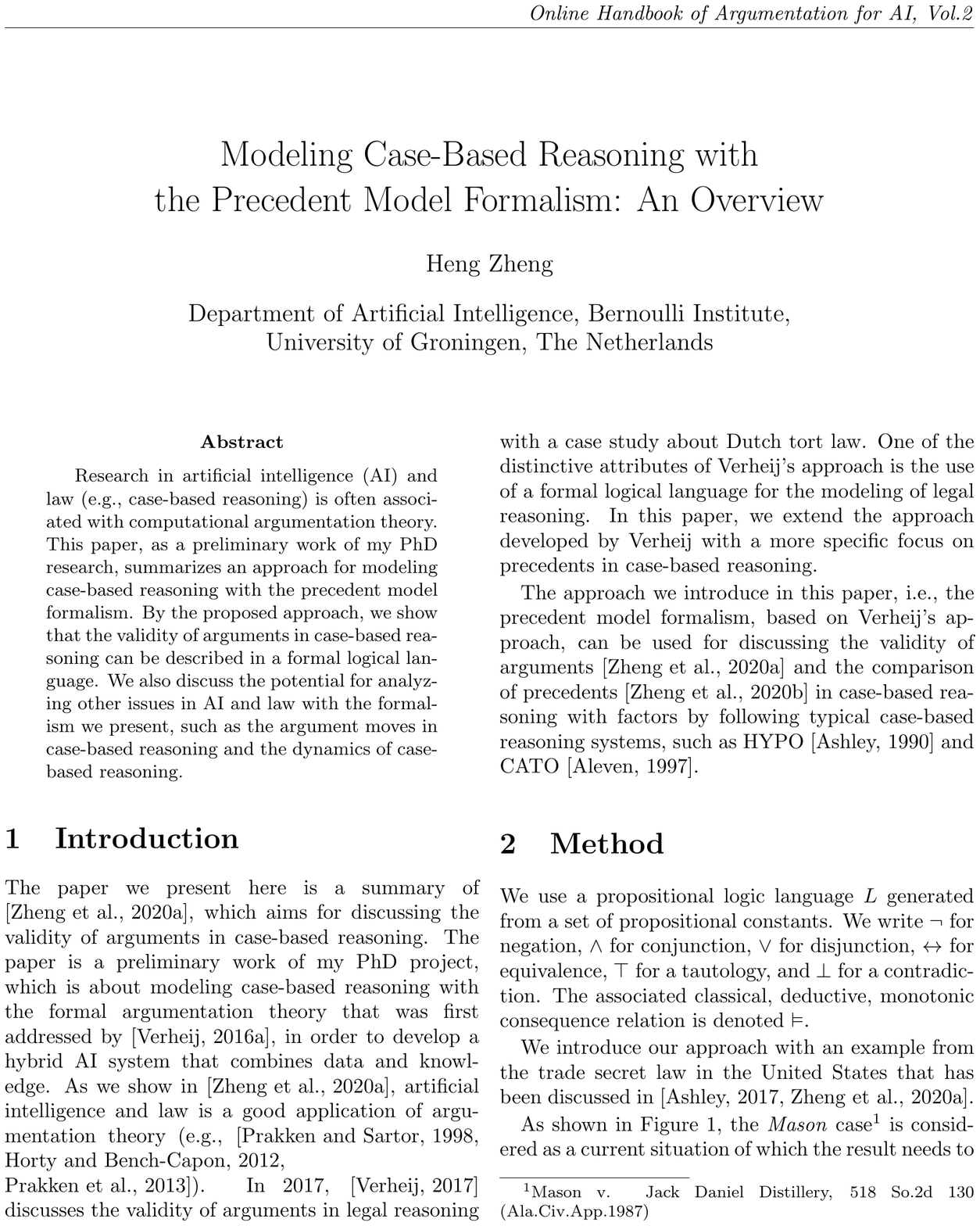}

\end{document}